\documentclass[conference]{IEEEtran}
\IEEEoverridecommandlockouts

\usepackage{cite}
\usepackage{amsmath,amssymb,amsfonts}
\usepackage{algorithmic}
\usepackage{graphicx}
\usepackage{textcomp}
\usepackage{xcolor}
\def\BibTeX{{\rm B\kern-.05em{\sc i\kern-.025em b}\kern-.08em
    T\kern-.1667em\lower.7ex\hbox{E}\kern-.125emX}}

\usepackage[linesnumbered,ruled,vlined]{algorithm2e}
\usepackage{epstopdf}
\usepackage{epsfig} 
\usepackage{times} 
\usepackage{amsmath, amssymb, amsfonts} 
\usepackage{float}
\usepackage{booktabs}
\usepackage{multirow}
\usepackage{utfsym} 
\usepackage[utf8]{inputenc}
\usepackage{pifont}

\usepackage{hyperref}
\hypersetup{
  colorlinks=true,         
  linkcolor=blue, 
  citecolor=black,
  filecolor=magenta,
  urlcolor=black  
}
\usepackage{bbm}

\usepackage{etoolbox}
\makeatletter
\patchcmd{\@IEEEpubidpullup}{\vskip\z@}{\vskip 1cm}{}{}
\newcommand{\linebreakand}{%
\end{@IEEEauthorhalign}
\hfill\mbox{}\par
\mbox{}\hfill\begin{@IEEEauthorhalign}
}

\begin{document}

\title{Prototype-Guided Non-Exemplar Continual Learning for Cross-subject EEG Decoding
\\

\thanks{This work was supported by the National Research Foundation of Korea (NRF) grant funded by the Korea government (MSIT) (No.2022-2-00975, MetaSkin: Developing Next-generation Neurohaptic Interface Technology that enables Communication and Control in Metaverse by Skin Touch) and the Institute of Information \& Communications Technology Planning \& Evaluation (IITP) grant, funded by the MSIT (No. RS-2019-II190079, Artificial Intelligence Graduate School Program (Korea University)).}
}

\IEEEpubid{%
    \raisebox{-1cm}{%
        \makebox[\columnwidth]{\fontsize{10}{12}\selectfont\fontfamily{ptm}\selectfont%
        979-8-3315-7927-2/26/\$31.00~\copyright~2026 IEEE\hfill}%
        \hspace{\columnsep}\makebox[\columnwidth]{}
    }
}

\author{
\IEEEauthorblockN{Dan Li}
\IEEEauthorblockA{\textit{Dept. of Artificial Intelligence} \\
\textit{Korea University}\\
Seoul, Republic of Korea \\
dan\_li@korea.ac.kr}
\and
\IEEEauthorblockN{Hye-Bin Shin}
\IEEEauthorblockA{\textit{Dept. of Brain and Cognitive Engineering} \\
\textit{Korea University}\\
Seoul, Republic of Korea \\
hb\_shin@korea.ac.kr}
\and
\IEEEauthorblockN{Yeon-Woo Choi}
\IEEEauthorblockA{\textit{Dept. of Artificial Intelligence} \\
\textit{Korea University}\\
Seoul, Republic of Korea \\
yw\_choi@korea.ac.kr}
}

\maketitle
\begin{abstract}
Due to the significant variability in electroencephalogram (EEG) signals across individuals, knowledge acquired from previous subjects is often overwritten as new subjects are introduced in continual EEG decoding tasks. Existing methods mainly rely on storing historical data from seen subjects as replay buffers to mitigate forgetting, which is impractical under privacy or memory constraints.
To address this issue, we propose a Prototype-guided Non-Exemplar Continual Learning (ProNECL) framework that preserves prior knowledge without accessing historical EEG samples.
ProNECL summarizes subject-specific discriminative representations into class-level prototypes and incrementally aligns new subject representations with a global prototype memory through prototype-based feature regularization and cross-subject alignment.
Experiments on the BCI Competition IV 2a and 2b datasets demonstrate that ProNECL effectively balances knowledge retention and adaptability, achieving superior performance in cross-subject continual EEG decoding tasks.
\end{abstract}

\begin{IEEEkeywords}
brain-computer interface, continual learning, electroencephalogram, motor imagery;
\end{IEEEkeywords}

\section{INTRODUCTION}
Brain-computer interfaces (BCIs) have found widespread applications in medical rehabilitation, offering innovative solutions for patients with motor disorders or those recovering from strokes. These systems enable individuals to control external devices, such as robotic arms, through motor imagery (MI) \cite{prabhakar2020framework,mao2019brain,cho2021neurograsp}, or to express intentions via imagined speech without the need for vocalization \cite{garcia2023intra,suk2011subject,ding2013changes}. In addition, electroencephalogram (EEG) signals are increasingly used to detect mental states \cite{yu2019weighted,suk2014predicting, myrden2015effects,lee2020continuous}, such as irregular brain activity linked to emotions, making it possible to perform emotion analysis \cite{ma2022few,kim2015abstract}. 
Despite these advancements, EEG signals pose significant challenges due to their high variability across individuals and even within the same person over time, complicating efforts to achieve consistent and accurate decoding. Although transfer learning \cite{thinker,lee1996multiresolution} and domain adaptation \cite{she2023improved,lee2018deep} aim to mitigate these challenges, they depend heavily on large source datasets that are often impractical in medical domains due to privacy concerns, and remain vulnerable to catastrophic forgetting (CF) when new data are introduced \cite{mane2021fbcnet,lee1995multilayer,french1999catastrophic}.

In an ideal scenario, intelligent systems should be capable of acquiring new knowledge from sequential data streams while preserving previously learned information. This concept, known as incremental or continual learning, is vital in artificial intelligence research. To mitigate the issue of catastrophic forgetting, various strategies have been proposed, including regularization techniques \cite{lee2015motion,rosenfeld2018incremental,lee1997new}, network expansion methods \cite{liu2021adaptive,lee2003pattern}, and memory replay approaches \cite{li2024domain, xiao2023online,bulthoff2003biologically,lee1999integrated}, with memory replay gaining attention for its simplicity and effectiveness.
However, direct storage and replay of raw data face significant limitations when dealing with privacy-sensitive or high-dimensional continuous signals such as electroencephalograms (EEG). On one hand, inter-subject physiological variations lead to unstable sample distributions, making effective feature alignment challenging for fixed-memory-based replay. On the other hand, from both privacy and storage cost perspectives, retaining large historical samples is impractical and may violate data security requirements. Consequently, traditional sample-based replay strategies prove unsuitable for such tasks, necessitating an efficient alternative that maintains knowledge continuity without relying on raw data.

To enable continual EEG decoding without storing large amounts of historical data while addressing the challenge of catastrophic forgetting,
we propose a novel Prototype-guided Non-Exemplar Continual Learning (ProNECL) framework that preserves previously learned knowledge through prototype-based representations and cross-subject feature alignment.
Specifically, ProNECL constructs class-level prototypes as compact summaries of discriminative knowledge and regularizes the representations of newly introduced subjects by aligning them with a global prototype memory, thereby maintaining representational consistency across subjects.
Extensive experimental results validate the effectiveness of our framework in mitigating forgetting, demonstrating its superiority in continual MI-EEG classification tasks.

\begin{figure}
    \centering
    \includegraphics[width=1.0\linewidth]{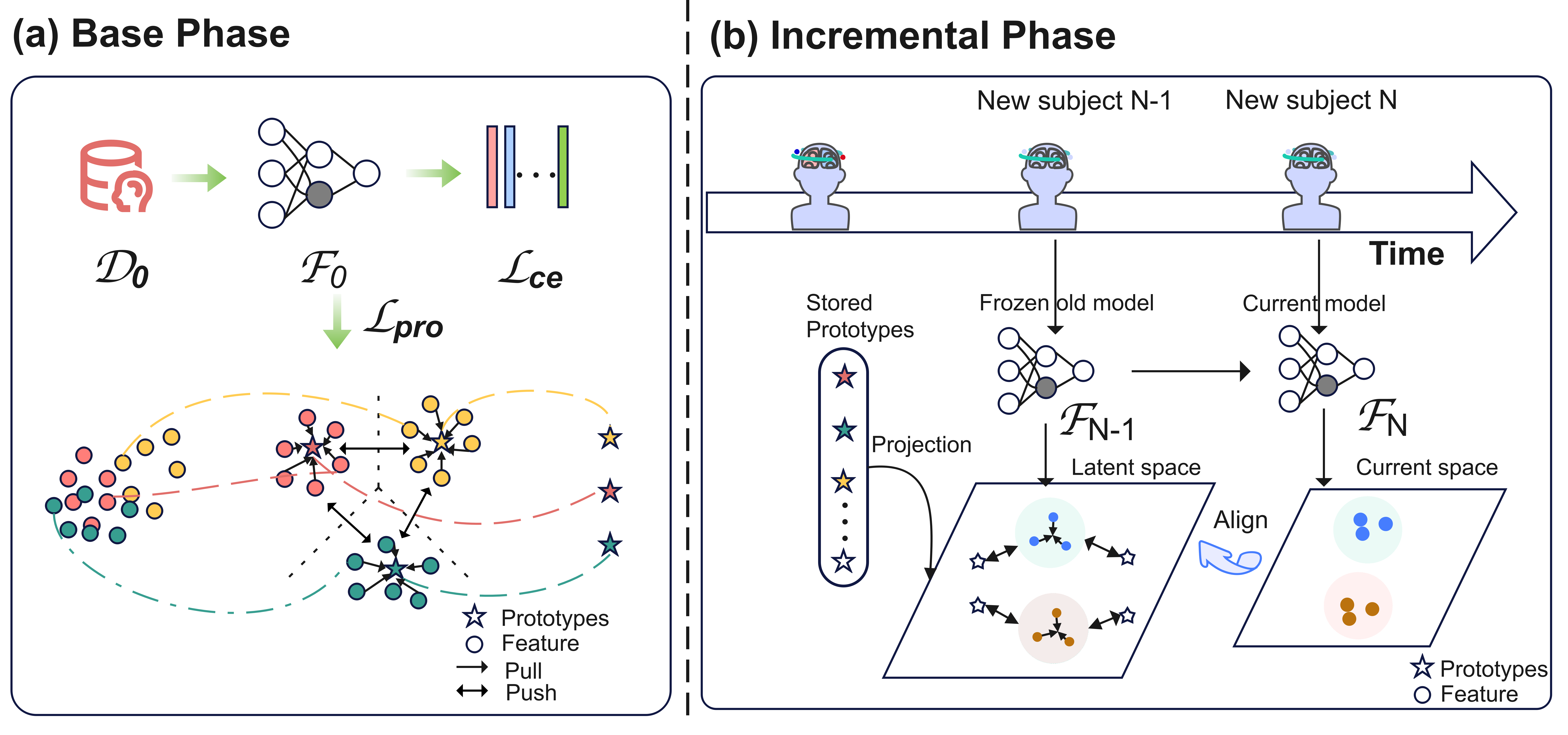}\vspace{-0.2cm}
    \caption{
Overview of ProNECL for continual EEG decoding. 
(1) \textbf{Base Phase:} A feature extractor $\mathcal{F}_0$ is first pre-trained on the initial dataset $\mathcal{D}_0$ to learn domain-invariant EEG representations. 
Class-level prototypes are then computed from $\mathcal{D}_0$ and stored as reference anchors in the prototype memory. 
(2) \textbf{Incremental Phase:}
When a new subject $\mathcal{D}_N$ arrives, the model $\mathcal{F}_N$ is trained by leveraging the previously learned prototype memory to regularize its latent representations, thereby preserving prior knowledge.
Specifically, class-level prototypes obtained from earlier subjects are projected into the latent space of $\mathcal{F}_N$ and used to align the feature distribution of the new subject with the established prototype space.
This prototype-guided representation alignment acts as a regularization mechanism that promotes cross-subject adaptation while maintaining consistency of learned representations, enabling effective knowledge retention without exemplar replay.
}
    \label{fig:model framework}
\end{figure}

\section{METHODOLOGY}
\subsection{Problem Definition}
EEG signals vary greatly across subjects, causing continual learning models to overfit new subjects and forget prior knowledge.
In real-world BCI applications, privacy and memory constraints often prohibit storing or replaying raw EEG data.
Thus, the objective is to train a model $\mathcal{F}: \mathcal{X} \rightarrow \mathcal{Y}$ capable of continual learning across subjects under a non-exemplar constraint.
Formally, let $\mathcal{V}={\mathcal{D}_1,\mathcal{D}_2,\ldots,\mathcal{D}_N}$ denote the sequential data stream of $N$ subjects, where $\mathcal{D}k={(X_k^i,Y_k^i,L_k^i){i=1}^{m_k}}$ represents the $k^\text{th}$ subject’s dataset with input $X_k\in\mathcal{X}$, class label $Y_k\in\mathcal{Y}$, domain label $L_k$, and $m_k$ samples.

\subsection{Feature Extraction and Prototype Construction}
For each subject $S_k$, the model $\mathcal{F}$ consists of an encoder $E_\phi$ and a classifier $C_\psi$. 
Given an EEG sample $X_k^i$, the encoder extracts the latent feature representation:
\begin{equation}
Z_k^i = E_\phi(X_k^i),
\end{equation}
where $Z_k^i \in \mathbb{R}^d$ denotes the $d$-dimensional embedding. 
The classifier $C_\psi$ then predicts the corresponding class label $\hat{Y}_k^i = C_\psi(Z_k^i)$ under supervised learning.

To summarize class-level information without storing raw samples, we introduce a prototype representation for each class $c \in \mathcal{Y}$. 
The prototype of class $c$ for the current subject $S_k$ is computed as the mean of all embeddings belonging to that class:
\begin{equation}
P_c^k = \frac{1}{|\mathcal{D}_c^k|} \sum_{(X_k^i, Y_k^i = c)} Z_k^i,
\end{equation}
where $\mathcal{D}_c^k$ denotes the subset of samples in $\mathcal{D}_k$ belonging to class $c$. After learning on $S_k$, the global prototype memory $\mathcal{P} = \{P_c\}_{c=1}^{C}$ is updated using an exponential moving average to integrate new subject information while maintaining prior knowledge:
\begin{equation}
P_c \leftarrow \alpha P_c + (1 - \alpha) P_c^k,
\end{equation}
where $\alpha \in [0,1]$ controls the balance between previously accumulated and newly acquired representations. Rather than relying on exemplar replay, our prototype representation abstracts each class as a compact summary of its learned distribution, enabling cross-subject adaptation without compromising privacy.

\subsection{Prototype-Guided Continual Learning}
Motivated by the goal of preserving knowledge across subjects without storing raw EEG data, we propose a prototype-guided learning strategy that aligns new subject representations with existing class prototypes.
During training on the $k^\text{th}$ subject, the model learns from $\mathcal{D}_k$ under a non-exemplar constraint, with no prior samples available.
To ensure stable retention, the objective combines supervised classification loss with a prototype-guided regularization term.

\paragraph{Supervised Classification Loss}
the classification head is optimized using the cross-entropy loss based solely on the current subject's labeled data:
\begin{equation}
\mathcal{L}_\text{ce} = 
-\mathbb{E}_{(x_k, y_k) \sim (X_k, Y_k)} 
\left[
\sum_{c=1}^{C} 
\mathbbm{1}_{[c = y_k]} 
\log \sigma(\mathcal{F}_k(x_k))_c
\right],
\end{equation}
where $C$ denotes the number of MI classes, and $\sigma$ is the softmax function applied to the classifier output.

\paragraph{Prototype Consistency and Cross-Subject Alignment}
to prevent the model from deviating from previously learned feature distributions, we introduce a prototype-guided consistency loss. 
For each sample $(x_k, y_k)$, the encoder output $E_\phi(x_k)$ is encouraged to stay close to its corresponding class prototype $P_{y_k}$ in the embedding space:
\begin{equation}
\mathcal{L}_\text{pro} = 
\mathbb{E}_{(x_k, y_k) \sim (X_k, Y_k)} 
\left[
\left\| 
E_\phi(x_k) - P_{y_k}
\right\|_2^2
\right].
\end{equation}
 In addition, to enhance cross-subject domain invariance, we align the mean embedding of the current subject with the global prototype centroid:
\begin{equation}
\mathcal{L}_\text{align} = 
\left\| 
\frac{1}{m_k} \sum_{i=1}^{m_k} E_\phi(X_k^i)
- 
\frac{1}{C} \sum_{c=1}^{C} P_c
\right\|_2^2,
\end{equation}
where the first term represents the subject-level mean feature of the current data, and the second term denotes the global centroid of all class prototypes. This alignment encourages the encoder to generate domain-invariant representations by pulling the current subject’s feature space toward the shared latent space, thereby mitigating inter-subject variability without relying on exemplar replay.

\begin{figure*}[t]
    \centering
    \includegraphics[width=0.65\linewidth]{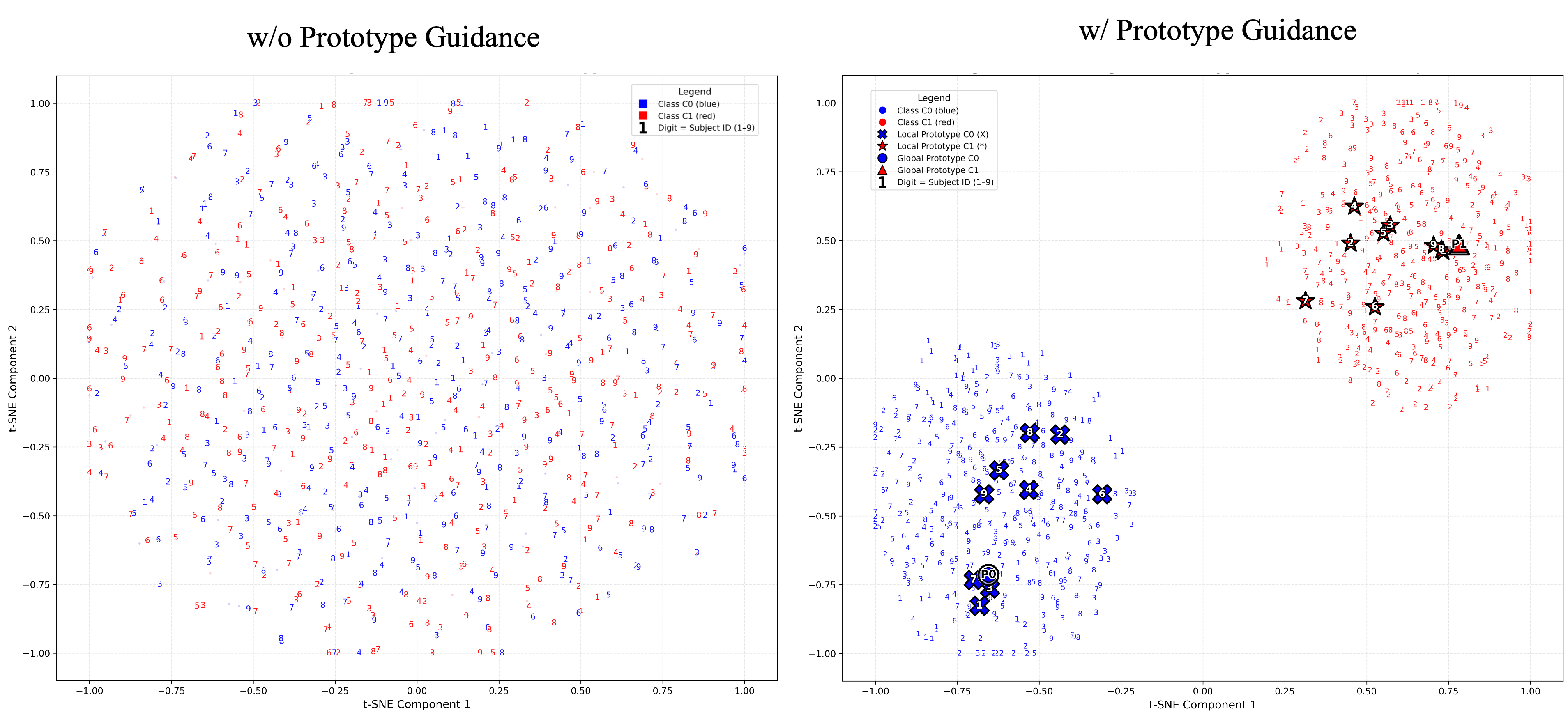}
    \caption{t-SNE of subject-invariant features on the 2a dataset (S1–S9): comparison without/with prototype guidance. Digits (1–9) indicate subject IDs, while markers “X”, “\ding{73}”, and “P” denote local and global prototypes, respectively.}
    \label{fig:proto-guidance}
\end{figure*} 

\paragraph{Overall Objective}
The total loss combines the above objectives:
\begin{equation}
\mathcal{L}_\text{total} = 
\mathcal{L}_\text{ce} + 
\lambda_\text{p} \mathcal{L}_\text{pro} + \lambda_\text{a} \mathcal{L}_\text{align},
\end{equation}
where $\lambda_\text{p}$ and $\lambda_\text{a}$ balance the prototype and alignment constraints.

\section{EXPERIMENTS}\label{sec:experiments}
\subsection{Datasets and Evaluation Metrics}
\label{sec:evaluation metrics}
In this study, we used the BCI Competition IV datasets 2a \cite{bcicomp2a} and 2b \cite{bcicomp2b} to evaluate our method on EEG data from nine subjects performing MI tasks. 
Dataset 2a includes four MI classes (left hand, right hand, foot, and tongue) recorded from 22 channels at 250 Hz, with 576 trials per subject across two sessions. 
Dataset 2b, containing two MI classes (left and right hand), was recorded from three channels at 250 Hz, comprising 720 trials per subject.
We measure a widely used metric backward transfer (BWT) \cite{lopez2017gradient} to assess the effect of new subject learning on previously seen subjects. BWT is calculated as: $\text{BWT} = \frac{1}{N-1} \sum_{i=1}^{N-1} \left( a_{N,i} - a_{i,i} \right)$, where \( a_{j,i} \) is the accuracy on subject \( i \) after training on subject \( j \). Negative BWT indicates forgetting, while positive BWT shows performance improvement on earlier subjects. We also calculate the average accuracy (ACC) across all subjects after the final round of learning to assess overall retention. 

\subsection{Baselines and Experimental Setting}
We compare our proposed ProNECL with representative continual learning methods, including subject-incremental EEG approaches and general non-exemplar algorithms adapted to domain-incremental settings.
\textbf{Finetuning:} Serves as the lower bound, sequentially training on each subject without forgetting mitigation.
\textbf{EWC \cite{kirkpatrick2017overcoming}:} Regularizes parameter updates using the Fisher information matrix to preserve important weights.
\textbf{MUDVI \cite{duan2024online}:} Utilizes a balanced memory buffer and temporal consistency for stable cross-subject learning.
\textbf{CGER \cite{deng2023centroid}:} Applies centroid-guided replay to align new and previous representations, reducing feature drift.
All methods share the same training setup using DeepConvNet (DCN) \cite{deepconvnet} as the feature extractor and a three-layer MLP as the domain classifier with ELU and softmax activations. The model is trained for 200 epochs with a learning rate of 0.001 and early stopping. In all experiments, we set $\lambda_\text{p}=0.1$ and $\lambda_\text{a}=0.3$ to balance the prototype and alignment constraints.

\begin{table}[t]
\caption{Performance comparison of ProNECL and baselines on two BCI benchmarks, with results reported as average accuracy (ACC, $\%$) and backward transfer (BWT, $\%$) with values representing the mean and standard deviation over five runs, and the best performance in bold.}
\label{tab:baseline_results}
\begin{center}
\resizebox{0.95\linewidth}{!}{
\begin{tabular}{lcccc}
\toprule
\multirow{4}{*}{Method} & \multicolumn{2}{c}{BCI-C IV 2a \cite{bcicomp2a}} & \multicolumn{2}{c}{BCI-C IV 2b \cite{bcicomp2b}} \\
\cmidrule(lr){2-3} \cmidrule(lr){4-5}
& ACC (std.) & BWT (std.) & ACC (std.) & BWT (std.) \\
\midrule
\multirow{1}{*}{Finetuning}
   & 32.33 (4.19)*** & -42.70 (6.96)
   & 55.39 (3.47)*** & -22.19 (3.88) \\
\multirow{1}{*}{EWC \cite{kirkpatrick2017overcoming}}
   & 44.67 (2.19)*** & -34.11 (2.76)
   & 60.08 (1.94)*** & -21.65 (1.89) \\
\multirow{1}{*}{MUDVI \cite{duan2024online}}
 & 46.41 (1.03)*** & -18.11 (1.27)
 & 67.20 (5.41)*** & -9.49 (5.60) \\
\multirow{1}{*}{CGER \cite{deng2023centroid}}
   & 49.84 (3.75)*** & -21.38 (2.44)
   & 67.43 (2.98)*** & -9.05 (2.77) \\
\multirow{1}{*}{ProNECL (Ours)}
  & \textbf{77.18 (1.76)} & 0.12 (1.53)
  &\textbf{81.15 (2.11)} & 0.33 (0.79) \\
\bottomrule
\end{tabular}
}
\end{center}
\footnotesize{{$^*$ACC: average accuracy in $\%$, BWT: backward transfer in $\%$, std.: standard deviation.
\textit{Significance levels comparing each method to ProNECL (Ours):} $^{*}p < 0.05$, $^{***}p < 0.001$.}}
\end{table}

\subsection{Results and Discussion}
\subsubsection{Model Performance Compared with Baselines}
Table I presents the average classification accuracy of ProNECL and several state-of-the-art baselines on the BCI Competition IV 2a and 2b datasets. 
Compared with conventional continual learning approaches, ProNECL achieves consistently higher performance across all subjects, demonstrating its effectiveness in mitigating catastrophic forgetting under the non-exemplar constraint. 
While Finetuning shows severe degradation as new subjects are introduced, the regularization-based method EWC partially alleviates forgetting but still suffers from domain drift. 
In contrast, ProNECL leverages prototype-guided representation and cross-subject alignment to maintain both stability and adaptability, achieving a better trade-off between knowledge retention and new subject adaptation. 
These results validate that prototype-based guidance can serve as an effective surrogate for exemplar replay in continual EEG decoding.

\subsubsection{T-SNE Visualization with and without Prototype Guidance}
To further examine the effect of prototype guidance on feature representation, we visualize the learned embeddings using t-SNE for models trained with and without the prototype alignment module. 
As shown in Fig. 2, without prototype guidance, the latent features of different subjects exhibit large inter-subject variability, leading to overlapping or dispersed class boundaries. 
In contrast, the model trained with prototype guidance produces more compact and separable clusters, where samples of the same class from different subjects are well aligned in the shared latent space. 
This demonstrates that prototype-based alignment effectively enforces domain-invariant representations, facilitating cross-subject consistency and improving generalization in continual EEG decoding.

\section{CONCLUSIONS}
In this paper, we proposed ProNECL, a Prototype-guided Non-Exemplar Continual Learning framework for cross-subject EEG decoding.
Unlike conventional replay-based approaches, ProNECL eliminates the dependency on storing historical EEG data, thereby addressing both privacy and memory constraints in practical BCI systems.
By constructing class-level prototypes and aligning subject-specific representations with a global prototype space, the proposed framework effectively mitigates catastrophic forgetting while preserving cross-subject representational consistency.
Moreover, the prototype-guided regularization mechanism promotes stability of learned representations across incremental updates without relying on exemplar replay.
Extensive experiments on the BCI Competition IV 2a and 2b datasets demonstrate that ProNECL achieves superior performance in continual motor imagery EEG classification, outperforming existing methods in terms of both knowledge retention and generalization.
In future work, we plan to extend this framework to multimodal and unsupervised continual decoding scenarios to further enhance adaptability in real-world BCI applications.


\bibliographystyle{IEEEtran}


\bibliography{Reference}

\end{document}